# C-Score: Beyond Accuracy for Robustness Assessment in Semi-Supervised Learning under Open-World Unlabeled Contamination

Tsao-Lun Chen, Chi-Cheng Fu, Han-Yi E. Chou, Shun-Feng Su, Fellow, IEEE

*Abstract*— **Pseudo-label-based semi-supervised learning (SSL) has achieved strong performance due to its simplicity and scalability. However, it is typically developed under a closed-world assumption that unlabeled data are drawn from the same distribution as labeled data. In practical deployment, unlabeled data are often collected from open environments and may contain out-of-distribution (OOD) samples. Under such contamination, OOD samples may still receive high-confidence predictions and be incorporated into training as if they were valid target examples. This creates an important evaluation problem: clean in-distribution test accuracy may appear stable even when the internal learning dynamics of SSL have already deteriorated. To address this issue, we study hidden collapse in pseudo-label-based SSL under open-world unlabeled contamination from a diagnostic evaluation perspective. We present C-Score, a compact framework that evaluates training behavior in three complementary spaces: prediction, feature representation, and optimization. C-Score includes Pseudo-label Entropy (PLE) and Class Concentration Index (CCI) for unlabeled prediction behavior, Semantic Drift (Sem-Drift) for deviation from labeled semantic anchors, and Gradient Alignment (Grad-Align) for the compatibility between labeled and unlabeled optimization. OOD Filtration Failure (OOD-FF) is further reported as an oracle metric for controlled analysis. Experiments on CIFAR-10 and CIFAR-100 with multiple OOD sources, varying contamination ratios, and four pseudo-label-based SSL algorithms show that C-Score metrics reveal hidden degradation that clean accuracy alone fails to detect: under SVHN contamination, CCI rises over 280% while best-accuracy remains within 3% of the uncontaminated baseline; near-OOD sources (CIFAR-100, STL-10) cause up to 14.9% accuracy collapse (FlexMatch, r=0.5), whereas far-OOD sources (Textures) are largely suppressed by the confidence threshold with no measurable degradation. The results suggest that clean accuracy alone is insufficient for evaluating SSL robustness in open-world environments, and that internal diagnostic signals are necessary for more reliable robustness assessment under unlabeled contamination.**

## I. Introduction

Semi-supervised learning (SSL) aims to learn from a small labeled set together with a much larger unlabeled pool [1]. Among existing SSL approaches, pseudo-label-based methods are particularly attractive due to their simplicity and scalability [2]. Most such methods, however, are developed under the assumption that labeled and unlabeled samples are drawn from the same distribution. This assumption may be violated in real-world deployment, where unlabeled data are often collected from open environments and may contain out-of-distribution (OOD) samples [3].

This mismatch is particularly problematic for pseudo-label-based SSL because unlabeled supervision is generated by the model's own predictions. When OOD samples receive high-confidence predictions, they may be incorporated into training as if they were valid target examples, even though they do not belong to the target label space. Existing work has studied robust SSL, open-world SSL, and unlabeled data quality control, but standard evaluation still largely relies on clean in-distribution test accuracy [4], [5]. As a result, internal degradation may remain hidden even when end-task accuracy appears stable. We refer to this evaluation blind spot as *accuracy masking*.

To study this problem, we present C-Score, a compact diagnostic framework for evaluating pseudo-label-based SSL under open-world unlabeled contamination. C-Score monitors training behavior from three complementary perspectives: prediction, feature representation, and optimization, through Pseudo-label Entropy (PLE), Class Concentration Index (CCI), Semantic Drift (Sem-Drift), and Gradient Alignment (Grad-Align), with OOD Filtration Failure (OOD-FF) reported as an oracle metric for controlled analysis. Experiments on CIFAR-10 and CIFAR-100 with multiple OOD sources and four pseudo-label-based SSL algorithms show that these signals reveal hidden degradation that clean test accuracy alone may fail to detect.

The contributions of this paper are threefold:

- We identify accuracy masking as an evaluation blind spot in pseudo-label-based SSL under open-world unlabeled contamination.
- We present C-Score, a diagnostic framework that assesses training behavior from prediction, feature, and optimization perspectives.
- We empirically show that these signals reveal hidden degradation that may not be visible from clean test accuracy alone.

Tsao-Lun Chen is with the Graduate Institute of Oral Biology, National Taiwan University College of Medicine, Taipei, Taiwan (e-mail: R14450024@ntu.edu.tw).

Chi-Cheng Fu is with NVIDIA, Santa Clara, CA, USA (e-mail: chichengf@nvidia.com).

Han-Yi E. Chou is with the Graduate Institute of Oral Biology, National Taiwan University College of Medicine, Taipei, Taiwan (e-mail: hyechou@ntu.edu.tw).

Shun-Feng Su, Fellow, IEEE, is with the Department of Electrical Engineering, National Taiwan University of Science and Technology, Taipei, Taiwan (e-mail: sfsu@mail.ntust.edu.tw.)

Shun-Feng Su and Han-Yi E. Chou are the co-corresponding authors of this article.

## II. Problem Formulation

### A. Pseudo-Label-Based Semi-Supervised Learning

Semi-supervised learning (SSL) addresses a classification setting where only a small labeled dataset is available, while a much larger unlabeled dataset can be collected at relatively low cost. Let the labeled set be denoted by $D_l = \{(x_i, y_i)\}_{i=1}^{N_l}$ and the unlabeled set by $D_u = \{u_j\}_{j=1}^{N_u}$, where $N_l \ll N_u$. Given an input space $\mathcal{X} \subseteq \mathbb{R}^d$ and a label space $\mathcal{Y} = \{1, \dots, K\}$, the objective of SSL is to learn a classifier $f_\theta: \mathcal{X} \longrightarrow \mathcal{Y}$ that generalizes well on the target task by utilizing both labeled and unlabeled data.

In this paper, we focus on pseudo-label-based SSL, where unlabeled samples contribute to training through targets derived from the model's own predictions. For an unlabeled sample $u$, the model produces a predictive distribution $p_\theta(y|u)$, from which a hard pseudo-label or a soft supervisory target is constructed. In many practical methods, this process is further coupled with confidence-based selection or weighting, so that the influence of unlabeled data depends directly on prediction confidence.

Such a design is effective when unlabeled samples are drawn from the same target distribution as labeled data. However, its reliability depends on the assumption that confident predictions are also semantically valid. When this assumption no longer holds, prediction confidence can cease to be a reliable proxy for supervision quality, and the resulting unlabeled training signal may become systematically misleading.

### B. Open-World Unlabeled Contamination

In real-world settings, unlabeled data are often collected from open environments rather than from a clean in-distribution source. As a result, the unlabeled pool may contain out-of-distribution (OOD) samples that do not belong to the target label space. Under this condition, the standard SSL assumption that labeled and unlabeled data are drawn from the same distribution no longer holds.

This setting is particularly challenging for pseudo-label-based SSL. Because unlabeled supervision is constructed from the model's own predictions, a confident prediction on an unlabeled sample is treated as meaningful supervision. In the presence of OOD contamination, however, prediction confidence no longer guarantees that the induced supervision is semantically valid. OOD samples may still receive confident predictions and enter training in the same way as in-distribution samples.

In this paper, we study this problem as open-world unlabeled contamination, where the unlabeled pool contains both in-distribution and OOD samples, but the model does not know their true identities during training. We further distinguish between near-OOD samples, which are semantically closer to the target distribution, and far-OOD samples, which are more clearly unrelated to the target task. In this work, we adopt the taxonomy used in OpenOOD [6]. This distinction is important because the two types of OOD can affect pseudo-label-based learning in different ways.

### C. Accuracy Masking

Standard SSL is typically evaluated by classification accuracy on a clean in-distribution test set. This evaluation is appropriate when the unlabeled pool is consistent with the target distribution. Under open-world unlabeled contamination, however, clean test accuracy may no longer provide a sufficient view of model reliability.

This limitation is especially important in pseudo-label-based SSL, where unlabeled supervision is generated by the model itself. When OOD samples receive confident predictions and are treated as valid training targets, contaminated supervision can alter prediction behavior, feature organization, and optimization compatibility before its effect becomes visible in final clean accuracy. As a result, a model may appear stable under conventional evaluation even though its internal learning process has already deteriorated.

We refer to this evaluation blind spot as accuracy masking. Let $f_{\text{clean}}$ denote a model trained with a clean unlabeled pool, and let $f_{\text{cont}}$ denote the same model trained under open-world unlabeled contamination. Let $\text{Acc}(f)$ denote clean test accuracy, and let $S(f)$ denote internal learning stability, reflecting prediction behavior, feature organization, and optimization compatibility. Accuracy masking occurs when

$$\text{Acc}(f_{\text{cont}}) \approx \text{Acc}(f_{\text{clean}}), \quad S(f_{\text{cont}}) < S(f_{\text{clean}}). \tag{1}$$

In other words, observable task accuracy remains similar, while the underlying learning dynamics have already degraded. This motivates the need for diagnostic signals beyond accuracy alone, which we develop in the next section.

## III. Diagnostic Framework

To expose hidden degradation under open-world unlabeled contamination, we introduce C-Score, a compact diagnostic framework that monitors collapse signals in three complementary spaces: prediction, feature, and optimization. The key premise is that clean test accuracy may remain stable even when pseudo-labeling dynamics and internal representations have already deteriorated. C-Score is designed to diagnose hidden degradation that may not appear in clean accuracy by examining three complementary aspects of unlabeled learning: whether pseudo-label predictions remain well-behaved, whether unlabeled representations stay aligned with labeled class structure, and whether unlabeled optimization remains compatible with supervised learning.

C-Score contains two levels of analysis. Macro-level metrics characterize the global behavior of the unlabeled prediction stream itself and require no labeled reference. Micro-level metrics explicitly compare unlabeled behavior against labeled anchors or supervised optimization signals. Under this formulation, Pseudo-label Entropy (PLE) and Class Concentration Index (CCI) are macro-level prediction-space metrics, whereas Semantic Drift (Sem-Drift) and Gradient Alignment (Grad-Align) are micro-level feature- and optimization-space metrics, respectively. We additionally report OOD Filtration Failure (OOD-FF) as an oracle-only metric for post-hoc analysis. The definitions below follow the actual implementation of our diagnostic code.

## A. Prediction Space Metrics: PLE and CCI

We first define two macro-level metrics to characterize the prediction behavior of unlabeled samples. Let $p(c|u_i)$ denote the model predictive distribution for unlabeled sample $u_i$, where $c \in \{1, \dots, K\}$ indexes the target classes and $K$ is the number of target classes.

**Pseudo-label Entropy (PLE):** PLE measures the average entropy of the unlabeled predictive distribution:

$$\text{PLE} = -\frac{1}{N}\sum_{i=1}^{N}\sum_{c=1}^{K} p(c|u_i)\log(p(c|u_i) + \epsilon), \quad (2)$$

where $N$ is the number of unlabeled samples in the batch. In our implementation, PLE is computed directly from the softmax probabilities of unlabeled samples, without threshold filtering. A low PLE indicates sharp predictions, whereas a high PLE indicates diffuse and uncertain predictions.

**Class Concentration Index (CCI):** To quantify whether unlabeled predictions are concentrated on a few classes, we first compute the average predicted class distribution over the unlabeled batch:

$$\overline{p_c} = \frac{1}{N}\sum_{i=1}^{N} p(c|u_i). \quad (3)$$

CCI is then defined as the KL divergence from this average distribution to the uniform distribution:

$$\text{CCI} = \sum_{c=1}^{K} \overline{p_c}\log\left(\frac{\overline{p_c}}{1/K}\right). \quad (4)$$

This definition follows the implementation, which uses the batch-mean soft class distribution rather than hard pseudo-label counts. A small CCI indicates balanced class usage, whereas a large CCI indicates that predictions are increasingly concentrated into a small subset of classes. Under contamination, such concentration suggests the formation of class-wise sinks.

## B. Feature Space Metric: Semantic Drift

Prediction-space statistics do not determine whether unlabeled representations remain aligned with the supervised class structure. A model may still produce sharp predictions while the associated unlabeled representations drift away from labeled anchors.

**Semantic Drift (Sem-Drift):** Sem-Drift is computed in the logit space. Let $z_x$ denote the logits of labeled samples, $z_u$ the logits of weakly augmented unlabeled samples, $y_i$ the ground-truth label of labeled sample $x_i$ and $\widehat{y_j}$ the hard pseudo-label of unlabeled sample $u_j$. Let $m_j \in \{0,1\}$ denote the confidence mask indicating whether $u_j$ passes the threshold. For each class $c$, the labeled centroid is

$$\mu_L^{(c)} = \frac{1}{|\{i: y_i = c\}|}\sum_{i:y_i=c} z_{x,i}, \quad (5)$$

and the masked pseudo-labeled unlabeled centroid is

$$\mu_U^{(c)} = \frac{1}{\left|\{j: \widehat{y_j} = c, m_j = 1\}\right|}\sum_{j:\widehat{y_j}=c,m_j=1} z_{u,j}. \quad (6)$$

The class-wise semantic drift is then

$$\text{Sem-Drift}^{(c)} = \left\|\mu_L^{(c)} - \mu_U^{(c)}\right\|_2, \quad (7)$$

and the overall metric is the average over classes for which both centroids are defined:

$$\text{Sem-Drift} = \frac{1}{\left|C_t^{\text{valid}}\right|}\sum_{c\in C_t^{\text{valid}}} \text{Sem-Drift}_t^{(c)}. \quad (8)$$

This matches the implementation, which computes per-class L2 distances between labeled logit centroids and masked pseudo-labeled weak-logit centroids, then averages over valid classes. A small Sem-Drift indicates that unlabeled samples assigned to a class remain close to the corresponding labeled anchor, whereas a large Sem-Drift indicates feature-space corruption.

## C. Optimization Space Metric: Gradient Alignment

Even when prediction distributions appear stable and class centroids remain structured, unlabeled supervision may still induce optimization signals that are incompatible with supervised learning. Motivated by gradient-based data selection in LLM training, such as GREATS [20], we use Grad-Align as a lightweight probe of supervised-unlabeled gradient compatibility.

**Gradient Alignment (Grad-Align):** This metric measures the compatibility between the optimization signal induced by unlabeled supervision and the supervised objective.

Let $L_x$ denote the labeled loss, $L_u$ the unlabeled loss, and $W_{fc}$ the parameters of the final fully connected classifier layer. We compute

$$g_x = \nabla_{W_{fc}} L_x; \quad g_u = \nabla_{W_{fc}} L_u. \quad (9)$$

Grad-Align is defined as the cosine similarity between the flattened gradients:

$$\text{Grad-Align} = \frac{g_x^\top g_u}{\|g_x\|_2\|g_u\|_2 + \epsilon}. \quad (10)$$

This definition follows the implementation, which extracts gradients of the labeled and unlabeled losses with respect to the final FC-layer parameters and computes their cosine similarity. A positive value indicates that unlabeled optimization reinforces supervised learning, a near-zero value indicates weak coupling, and a negative value indicates direct gradient conflict.

## D. Oracle Metric: OOD Filtration Failure

For controlled analysis, we additionally report an oracle-only metric to verify whether OOD contamination is actually entering training through the confidence filter.

**OOD Filtration Failure (OOD-FF):** This metric quantifies how often OOD samples pass the confidence mask relative to ID samples, and therefore requires ground-truth ID/OOD partition labels.

Let $r_{ID}$ and $r_{OOD}$ denote the fractions of ID and OOD unlabeled samples that pass the confidence mask, respectively. OOD-FF is defined as

$$\text{OOD-FF} = \frac{r_{OOD}}{r_{ID} + \epsilon}. \quad (11)$$

This follows the implementation exactly. A small value indicates effective suppression of OOD samples, whereas a large value indicates that the masking mechanism is failing to reject OOD contamination relative to ID samples.

### E. Summary

C-Score provides a compact diagnosis of SSL failure under open-world contamination. PLE and CCI characterize the uncertainty and class concentration of unlabeled predictions, Sem-Drift measures deviation from labeled semantic anchors in logit space, and Grad-Align quantifies whether unlabeled optimization remains consistent with supervised learning. OOD-FF is used only as an oracle metric to verify contamination leakage. These signals expose hidden collapse beyond clean test accuracy.

## IV. Experimental Results

### A. Experimental Setup

**Datasets and OOD Taxonomy:** CIFAR-10 [7] was used as the primary in-distribution (ID) task, with 40 labeled samples in total and 50,000 unlabeled ID images. CIFAR-100 [7] was used as a harder 100-class setting, using 400 labeled samples and the full 50,000-image training set as the unlabeled ID pool. For CIFAR-10 contamination, we consider six OOD sources: CIFAR-100, STL-10 [8], SVHN [9], MNIST [10], Gaussian Noise, and Textures [11]. Following OpenOOD [12], CIFAR-100 and STL-10 were treated as near-OOD, while SVHN, MNIST, Gaussian Noise, and Textures were treated as far-OOD. To validate this taxonomy, dataset class names were encoded with the CLIP text encoder [13], and their pairwise cosine dissimilarities were visualized using Multidimensional Scaling (MDS) in Fig. 1.

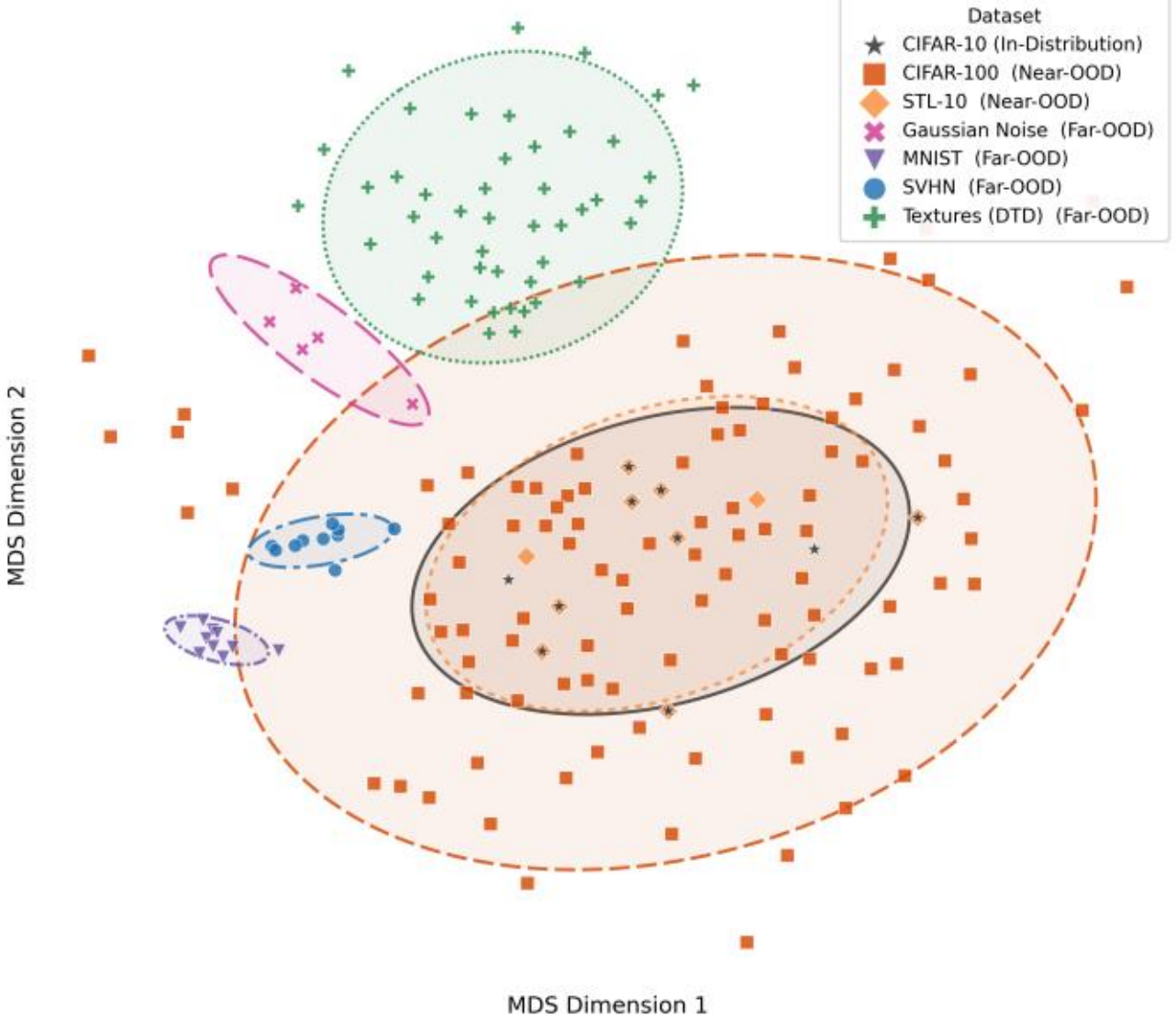

Fig. 1. MDS projection of CLIP text embeddings for class labels from each dataset. Spatial distance reflects cosine dissimilarity in semantic embedding space. Near-OOD datasets (CIFAR-100 and STL-10) overlap with CIFAR-10, whereas far-OOD datasets (SVHN, MNIST, Gaussian Noise, and DTD Textures) are semantically distant.

**OOD Contamination Protocol:** Following RE-SSL [14], the ID unlabeled pool was kept fixed and only the number of OOD samples was varied, in order to satisfy the controlled-variable principle. The contamination level was defined as:

$$r = \frac{N_{OOD}}{N_{ID} + N_{OOD}}, \quad (12)$$

where $N_{ID} = 50{,}000$ and $r \in \{0.1, 0.2, 0.3, 0.4, 0.5\}$. This design avoids the confounding effect of prior evaluations that fixed the total unlabeled set size while changing the unseen-class proportion, which simultaneously reduced the amount of ID unlabeled data. For DTD Textures, only the $r = 0.1$ setting was valid because the dataset contains 5,640 images; higher-ratio settings were excluded.

**Compared SSL Algorithms:** Four pseudo-label-based SSL methods were evaluated: FixMatch [15], FlexMatch [16], SoftMatch [17], and DS3L [18]. These methods were chosen to provide a representative comparison across major pseudo-label control mechanisms for unlabeled supervision. FixMatch serves as the baseline with a standard fixed-threshold design; FlexMatch extends this paradigm with curriculum-based class-adaptive thresholds; SoftMatch replaces hard selection with confidence-aware soft weighting and uniform alignment; and DS3L introduces similarity-based importance weighting using class centroids.

**Training, Implementation, and Evaluation Protocol:** All experiments used WRN-28-2 [19]. CIFAR-10 was trained for 200 epochs with 256 iterations per epoch, and CIFAR-100 for 500 epochs with 512 iterations per epoch. Each iteration used 64 labeled and 448 unlabeled samples. SGD was used with learning rate 0.03 and weight decay $5\times10^{-4}$, and EMA decay was set to 0.999. All results are reported as mean ± standard deviation over five random seeds, using the clean test accuracy of the EMA model as the primary metric. C-Score metrics (PLE, CCI, Sem-Drift, and Grad-Align) were logged during training, whereas OOD-FF was used only for oracle post-hoc analysis.

In total, the study comprised approximately 620 training runs across 4 SSL algorithms, 6 OOD sources, 5 contamination ratios, and 5 seeds. FixMatch was trained on the NVIDIA RTX PRO 6000 Blackwell server, while FlexMatch, SoftMatch, and DS3L were trained on NVIDIA L40S GPUs. V100 resources were used only for development and compatibility testing. The pipeline used NVIDIA cuDNN benchmark mode, AMP, torch.compile(), PyTorch native fused SGD, and NVIDIA DALI, yielding 2.64× speedup over a GPU-only baseline and 37.47× over CPU-only training. Code and ablation scripts are available at: https://github.com/wahahahaya/c-score.

### B. Main Evidence of Accuracy Masking

Table 1 summarizes the final-stage diagnostic states of the four SSL algorithms under clean training and severe OOD contamination. It shows that near-OOD sources generally induce larger accuracy drops, whereas far-OOD sources are more often suppressed by confidence thresholding; however, internal metrics, especially CCI, still reveal hidden degradation even when accuracy changes are limited.

Table 1. Summary of final-stage diagnostic metrics under clean training and the highest valid contamination setting for each OOD source. For CIFAR-100, STL-10, SVHN, MNIST, and Gaussian Noise, the severe contamination setting is $r = 0.5$. For Textures, only $r = 0.1$is valid due to limited sample availability.

| Algorithm | OOD Source | Acc (%) | PLE | CCI | Sem-Drift | Grad-Align | OOD-FF |
|---|---|---|---|---|---|---|---|
| **FixMatch** | clean ($r = 0.0$) | 71.34 ± 4.31 | 0.433 ± 0.242 | 0.143 ± 0.081 | 0.129 ± 0.028 | 0.000 ± 0.001 | 0.000 |
| | CIFAR-100 | 62.53 ± 2.11 | 0.689 ± 0.228 | 0.229 ± 0.149 | 0.139 ± 0.013 | -0.000 ± 0.000 | 0.022 ± 0.005 |
| | STL-10 | 63.36 ± 5.85 | 0.698 ± 0.131 | 0.151 ± 0.065 | 0.143 ± 0.006 | -0.001 ± 0.001 | 0.020 ± 0.004 |
| | SVHN | 70.51 ± 5.60 | 0.393 ± 0.331 | 0.552 ± 0.173 | 0.131 ± 0.011 | 0.000 ± 0.000 | 0.028 ± 0.018 |
| | MNIST | 67.84 ± 2.48 | 0.619 ± 0.410 | 0.495 ± 0.257 | 0.126 ± 0.010 | 0.000 ± 0.001 | 0.016 ± 0.020 |
| | Gaussian | 69.04 ± 3.49 | 0.560 ± 0.307 | 0.482 ± 0.231 | 0.125 ± 0.005 | 0.001 ± 0.000 | 0.011 ± 0.015 |
| | Textures | 73.32 ± 4.53 | 0.316 ± 0.127 | 0.178 ± 0.058 | 0.124 ± 0.018 | -0.000 ± 0.001 | 0.039 ± 0.006 |
| **FlexMatch** | clean ($r = 0.0$) | 79.19 ± 2.74 | 0.366 ± 0.172 | 0.018 ± 0.007 | 0.153 ± 0.027 | 0.002 ± 0.001 | 0.000 |
| | CIFAR-100 | 64.30 ± 1.79 | 0.856 ± 0.058 | 0.052 ± 0.012 | 0.218 ± 0.005 | 0.001 ± 0.000 | 0.028 ± 0.000 |
| | STL-10 | 68.40 ± 4.05 | 0.728 ± 0.280 | 0.059 ± 0.040 | 0.193 ± 0.034 | 0.000 ± 0.001 | 0.028 ± 0.002 |
| | SVHN | 69.43 ± 4.36 | 0.363 ± 0.128 | 0.285 ± 0.056 | 0.165 ± 0.007 | 0.001 ± 0.001 | 0.027 ± 0.002 |
| | MNIST | 70.51 ± 4.65 | 0.842 ± 0.315 | 0.224 ± 0.145 | 0.159 ± 0.006 | 0.001 ± 0.001 | 0.011 ± 0.004 |
| | Gaussian | 70.97 ± 5.75 | 0.624 ± 0.358 | 0.350 ± 0.188 | 0.167 ± 0.020 | 0.002 ± 0.001 | 0.012 ± 0.011 |
| | Textures | 76.96 ± 6.50 | 0.426 ± 0.142 | 0.025 ± 0.008 | 0.161 ± 0.024 | 0.001 ± 0.001 | 0.029 ± 0.002 |
| **SoftMatch** | clean ($r = 0.0$) | 72.65 ± 2.59 | 0.822 ± 0.165 | 0.018 ± 0.004 | 0.242 ± 0.025 | -0.000 ± 0.001 | 0.000 |
| | CIFAR-100 | 60.00 ± 3.80 | 1.077 ± 0.092 | 0.059 ± 0.016 | 0.265 ± 0.009 | 0.000 ± 0.000 | 0.031 |
| | STL-10 | 63.98 ± 5.71 | 1.081 ± 0.050 | 0.035 ± 0.014 | 0.262 ± 0.012 | 0.000 ± 0.001 | 0.031 |
| | SVHN | 68.75 ± 4.63 | 0.489 ± 0.184 | 0.276 ± 0.084 | 0.190 ± 0.014 | 0.000 ± 0.001 | 0.032 ± 0.001 |
| | MNIST | 68.28 ± 2.53 | 0.993 ± 0.172 | 0.138 ± 0.077 | 0.187 ± 0.011 | 0.000 ± 0.001 | 0.031 ± 0.001 |
| | Gaussian | 67.18 ± 3.95 | 0.981 ± 0.267 | 0.124 ± 0.123 | 0.183 ± 0.012 | 0.000 ± 0.001 | 0.031 ± 0.001 |
| | Textures | 73.18 ± 3.97 | 0.713 ± 0.207 | 0.024 ± 0.003 | 0.224 ± 0.028 | 0.000 ± 0.001 | 0.031 ± 0.001 |
| **DS3L** | clean ($r = 0.0$) | 72.48 ± 6.47 | 0.500 ± 0.212 | 0.121 ± 0.070 | 0.141 ± 0.014 | 0.001 ± 0.001 | 0.000 |
| | CIFAR-100 | 61.56 ± 3.50 | 0.703 ± 0.191 | 0.281 ± 0.117 | 0.145 ± 0.009 | -0.001 ± 0.001 | 0.024 ± 0.004 |
| | STL-10 | 62.02 ± 4.42 | 0.637 ± 0.234 | 0.271 ± 0.095 | 0.136 ± 0.012 | -0.001 ± 0.000 | 0.022 ± 0.005 |
| | SVHN | 70.17 ± 2.43 | 0.282 ± 0.198 | 0.613 ± 0.099 | 0.122 ± 0.006 | 0.001 ± 0.001 | 0.034 ± 0.015 |
| | MNIST | 65.22 ± 4.11 | 0.637 ± 0.318 | 0.472 ± 0.224 | 0.135 ± 0.009 | 0.001 ± 0.001 | 0.013 ± 0.016 |
| | Gaussian | 66.77 ± 2.67 | 0.775 ± 0.382 | 0.318 ± 0.254 | 0.128 ± 0.008 | 0.001 ± 0.001 | 0.008 ± 0.018 |
| | Textures | 73.33 ± 4.17 | 0.324 ± 0.164 | 0.127 ± 0.049 | 0.118 ± 0.020 | 0.001 ± 0.001 | 0.037 ± 0.004 |

### *C. Diagnostic Trends Across Contamination Levels*

Fig. 2 shows the training dynamics of the proposed diagnostics for FixMatch on CIFAR-10 with SVHN contamination. CCI exhibits the clearest and most stable separation across contamination levels, indicating progressively stronger class concentration under heavier OOD contamination. Sem-Drift also increases with $r$, but with weaker separation. PLE decreases early in training and increases later under heavier contamination, suggesting reduced stability of pseudo-label behavior. In contrast, Grad-Align remains close to zero and provides little separation, indicating that the observed degradation is more visible in prediction and representation spaces than in coarse final-layer gradient compatibility. Overall, CCI is the strongest signal, with Sem-Drift and PLE providing complementary evidence.

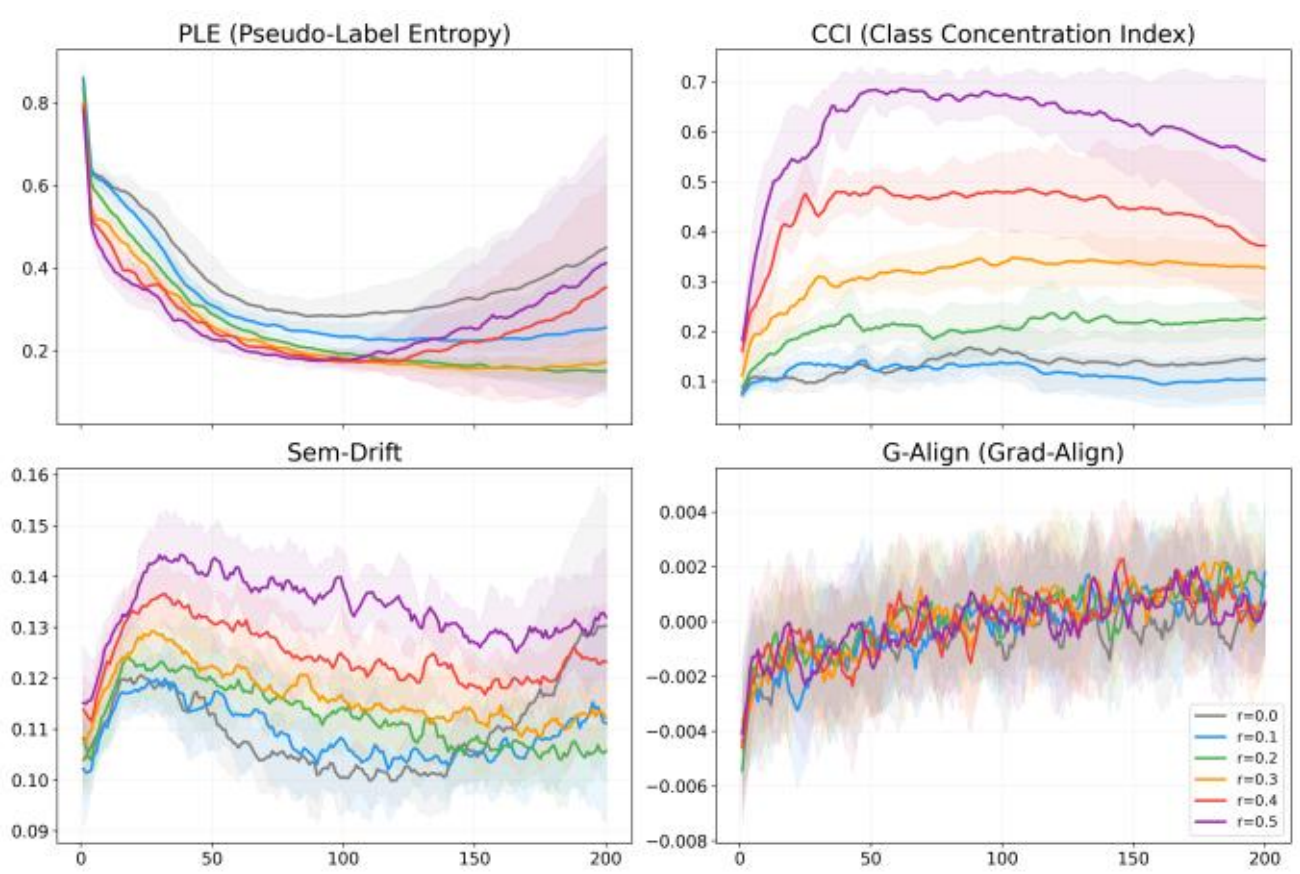

Fig. 2. Training dynamics of C-Score metrics for FixMatch on CIFAR-10 under SVHN contamination. CCI shows the clearest and most stable separation across contamination levels. Sem-Drift provides weaker but consistent complementary evidence, while PLE decreases early and rebounds later under heavier contamination. Grad-Align remains close to zero and shows limited discriminative power in this setting.

### *D. Comparison Across SSL Algorithms*

We next compared FixMatch, FlexMatch, SoftMatch, and DS3L on CIFAR-10 with SVHN contamination. Fig. 3(a) shows clean test accuracy versus contamination level $r$, and Fig. 3(b) shows the corresponding CCI values. Although the four methods exhibit different accuracy trajectories, their differences become clearer when internal diagnostic behavior is considered jointly with end-task performance.

FlexMatch achieves the strongest clean baseline accuracy, but its performance degrades as contamination increases. FixMatch shows a more moderate accuracy drop, while SoftMatch and DS3L remain in a similar range under heavier contamination. However, all four methods show increasing CCI with larger $r$, indicating progressively stronger class concentration under OOD contamination. FlexMatch and SoftMatch maintain lower absolute CCI than FixMatch and DS3L, whereas DS3L exhibits the sharpest increase at high contamination levels. These results suggest that algorithm ranking based on clean accuracy alone can be incomplete, since methods with comparable accuracy may still differ substantially in their internal stability.

Overall, the comparison shows that pseudo-label control mechanisms affect not only final performance but also the extent of hidden degradation under contamination. This further supports the need to evaluate SSL robustness using both accuracy and diagnostic signals.

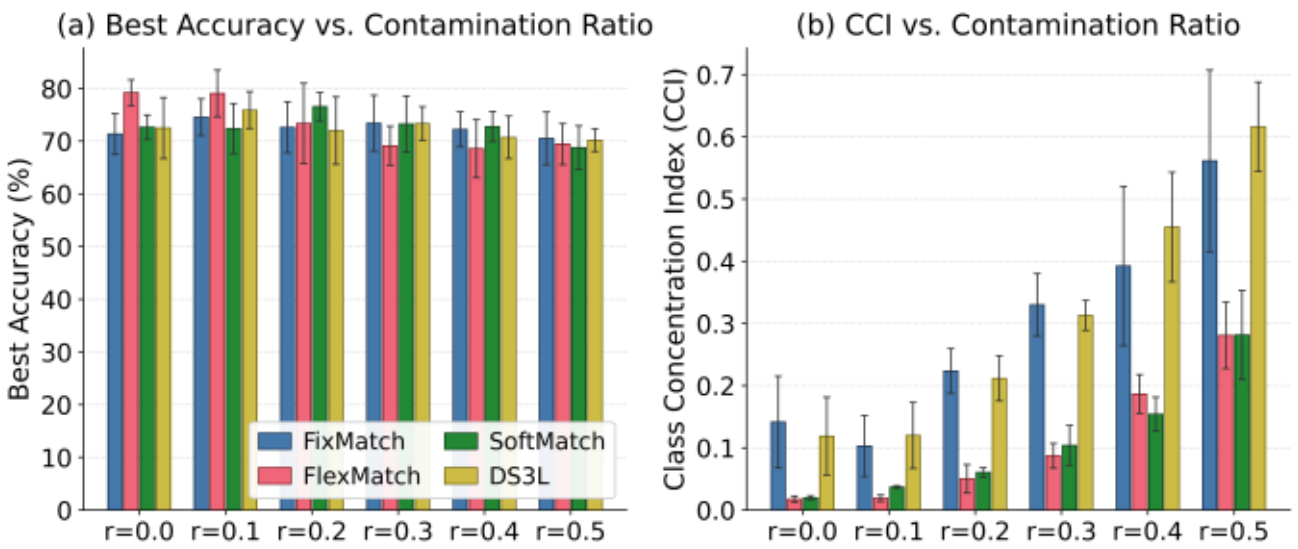

Fig. 3. Comparison across pseudo-label-based SSL algorithms on CIFAR-10 with SVHN contamination. (a) Clean test accuracy versus contamination level $r$. (b) Class Concentration Index (CCI) versus contamination level $r$. While accuracy differences across methods are moderate, CCI reveals clearer differences in internal degradation under increasing OOD contamination.

### E. Effect of OOD Source Type

OOD source type has a clear effect on robustness. Under FixMatch, near-OOD sources such as CIFAR-100 and STL-10 cause the largest accuracy degradation, whereas clearly unrelated sources such as Textures are largely suppressed by the confidence threshold. However, some far-OOD sources, including SVHN, MNIST, and Gaussian Noise, still produce strong increases in CCI despite only moderate accuracy loss. This shows that source type influences not only how much performance drops, but also whether degradation remains hidden from clean accuracy alone.

### F. Evaluation on CIFAR-100

To verify whether accuracy masking extends to harder classification settings, FixMatch was additionally evaluated with CIFAR-100 as the ID task under SVHN contamination. Table 2 summarizes the results across contamination ratios. Despite a 100-class target space, clean test accuracy remains largely stable (43.21% → 39.22%), while CCI rises sharply from 0.161 to 1.430, corresponding to relative increase of approximately 787%. This confirms that internal diagnostic degradation can outpace observable accuracy loss even in harder settings, reinforcing the diagnostic value of CCI beyond the CIFAR-10 baseline.

| *r* | Best Acc (%) | CCI |
|---|---|---|
| 0.0 | 43.21 ± 1.07 | 0.161 ± 0.015 |
| 0.1 | 43.47 ± 1.21 | 0.339 ± 0.026 |
| 0.2 | 43.51 ± 1.35 | 0.578 ± 0.072 |
| 0.3 | 42.47 ± 1.01 | 0.728 ± 0.124 |
| 0.4 | 41.63 ± 1.43 | 1.183 ± 0.106 |
| 0.5 | 39.22 ± 1.26 | 1.430 ± 0.238 |

## V. Conclusion

Clean test accuracy alone is insufficient for evaluating pseudo-label-based SSL under open-world unlabeled contamination. C-Score reveals hidden degradation from prediction, feature, and optimization perspectives, with CCI showing the most consistent sensitivity across settings. These results support moving beyond accuracy-only evaluation in contaminated SSL and suggest that C-Score may serve as a practical basis for future contamination-aware SSL design. Future work may extend C-Score from post-hoc diagnosis to training-time contamination detection and regularization.

## Acknowledgment

The authors gratefully acknowledge National Taiwan University of Science and Technology and NVIDIA for their support of this work. The computational resources used in this study included an NVIDIA RTX PRO 6000 Blackwell server and V100 GPUs provided by National Taiwan University of Science and Technology, as well as L40S GPU cloud computing resources provided through NVIDIA Brev. The authors also thank the collaborators and technical staff for their technical guidance and infrastructure assistance.

## Appendix: Extended Experimental Results

This appendix reports the full per-run and per-seed statistics underlying the summary results in Section V, aggregated as mean ± standard deviation over n = 5 random seeds (seeds 0, 10, 20, 30, 40). All accuracy figures are the best clean-test EMA accuracy achieved during training; all diagnostic metrics (PLE, CCI, Sem-Drift, Grad-Align, OOD-FF) are averaged over the final 10 training epochs, consistent with the aggregation protocol used to produce Tables I and II. In total, 276 individual training runs are summarized across the three tables below.

### *A. Full OOD-Source × Contamination-Ratio Sweep (FixMatch, CIFAR-10)*

Table A1 extends Table I by reporting all six diagnostic dimensions for FixMatch on CIFAR-10 across every OOD source and every contamination ratio r that was evaluated, rather than only the clean and worst-case rows. For DTD Textures, only r = 0.1 is reported; as noted in Section IV, the Textures corpus contains only 5,640 images, which is insufficient to support distinct sampling at higher contamination ratios without image reuse across runs, so r > 0.1 settings are excluded from this table.

TABLE A1
FixMatch on CIFAR-10: Full Metric Sweep by OOD Source and Contamination Ratio

| OOD Source | r | n | Acc. (%) | PLE | CCI | Sem-Drift | Grad-Align | OOD-FF |
|---|---|---|---|---|---|---|---|---|
| CIFAR-100 | 0.0 | 5 | 71.34±4.31 | 0.433±0.242 | 0.143±0.081 | 0.129±0.028 | 0.0005±0.0006 | 0.000±0.000 |
| CIFAR-100 | 0.1 | 5 | 70.95±3.40 | 0.552±0.181 | 0.094±0.051 | 0.134±0.012 | -0.0002±0.0005 | 0.014±0.003 |
| CIFAR-100 | 0.2 | 5 | 68.98±4.61 | 0.583±0.267 | 0.134±0.077 | 0.125±0.019 | 0.0006±0.0008 | 0.017±0.004 |
| CIFAR-100 | 0.3 | 5 | 66.99±2.92 | 0.625±0.217 | 0.149±0.084 | 0.136±0.016 | -0.0003±0.0003 | 0.018±0.004 |
| CIFAR-100 | 0.4 | 5 | 65.21±3.70 | 0.640±0.227 | 0.175±0.085 | 0.135±0.017 | 0.0001±0.0009 | 0.022±0.004 |
| CIFAR-100 | 0.5 | 5 | 62.53±2.11 | 0.689±0.228 | 0.229±0.149 | 0.139±0.013 | -0.0001±0.0005 | 0.022±0.005 |
| STL-10 | 0.0 | 5 | 71.34±4.31 | 0.433±0.242 | 0.143±0.081 | 0.129±0.028 | 0.0005±0.0006 | 0.000±0.000 |
| STL-10 | 0.1 | 5 | 71.11±6.04 | 0.526±0.112 | 0.127±0.045 | 0.136±0.010 | -0.0004±0.0006 | 0.023±0.003 |
| STL-10 | 0.2 | 5 | 68.66±4.16 | 0.520±0.181 | 0.173±0.068 | 0.133±0.011 | -0.0004±0.0005 | 0.025±0.002 |
| STL-10 | 0.3 | 5 | 67.00±3.86 | 0.597±0.162 | 0.165±0.085 | 0.142±0.012 | -0.0007±0.0008 | 0.023±0.004 |
| STL-10 | 0.4 | 5 | 64.60±4.67 | 0.625±0.188 | 0.220±0.121 | 0.138±0.010 | -0.0001±0.0006 | 0.023±0.004 |
| STL-10 | 0.5 | 5 | 63.36±5.85 | 0.698±0.131 | 0.151±0.065 | 0.143±0.006 | -0.0009±0.0009 | 0.020±0.004 |
| SVHN | 0.0 | 5 | 71.34±4.31 | 0.433±0.242 | 0.143±0.081 | 0.129±0.028 | 0.0005±0.0006 | 0.000±0.000 |
| SVHN | 0.1 | 5 | 74.55±3.90 | 0.252±0.182 | 0.104±0.055 | 0.113±0.020 | 0.0009±0.0012 | 0.042±0.009 |
| SVHN | 0.2 | 5 | 72.60±5.41 | 0.151±0.033 | 0.225±0.042 | 0.106±0.004 | 0.0015±0.0014 | 0.037±0.002 |
| SVHN | 0.3 | 5 | 73.40±5.95 | 0.169±0.065 | 0.329±0.061 | 0.113±0.004 | 0.0005±0.0010 | 0.037±0.006 |
| SVHN | 0.4 | 5 | 72.23±3.71 | 0.336±0.271 | 0.377±0.144 | 0.123±0.011 | 0.0003±0.0003 | 0.027±0.013 |
| SVHN | 0.5 | 5 | 70.51±5.60 | 0.393±0.331 | 0.552±0.173 | 0.131±0.011 | 0.0005±0.0004 | 0.028±0.018 |
| MNIST | 0.0 | 5 | 71.34±4.31 | 0.433±0.242 | 0.143±0.081 | 0.129±0.028 | 0.0005±0.0006 | 0.000±0.000 |
| MNIST | 0.1 | 5 | 75.85±6.30 | 0.264±0.158 | 0.123±0.056 | 0.109±0.021 | 0.0004±0.0012 | 0.045±0.010 |
| MNIST | 0.2 | 5 | 73.93±5.13 | 0.264±0.101 | 0.266±0.101 | 0.124±0.017 | 0.0006±0.0005 | 0.046±0.006 |
| MNIST | 0.3 | 5 | 73.52±4.33 | 0.261±0.088 | 0.338±0.099 | 0.124±0.011 | 0.0008±0.0013 | 0.040±0.008 |
| MNIST | 0.4 | 5 | 71.21±5.63 | 0.316±0.182 | 0.503±0.140 | 0.120±0.003 | 0.0010±0.0007 | 0.026±0.015 |
| MNIST | 0.5 | 5 | 67.84±2.48 | 0.619±0.410 | 0.495±0.257 | 0.126±0.010 | 0.0003±0.0008 | 0.016±0.020 |
| Gaussian Noise | 0.0 | 5 | 71.34±4.31 | 0.433±0.242 | 0.143±0.081 | 0.129±0.028 | 0.0005±0.0006 | 0.000±0.000 |
| Gaussian Noise | 0.1 | 5 | 75.63±6.58 | 0.159±0.024 | 0.103±0.042 | 0.097±0.003 | 0.0012±0.0005 | 0.040±0.001 |
| Gaussian Noise | 0.2 | 5 | 76.47±4.58 | 0.175±0.061 | 0.167±0.049 | 0.107±0.007 | 0.0013±0.0008 | 0.041±0.003 |
| Gaussian Noise | 0.3 | 5 | 71.27±5.37 | 0.139±0.009 | 0.373±0.064 | 0.113±0.009 | 0.0007±0.0005 | 0.041±0.000 |
| Gaussian Noise | 0.4 | 5 | 69.08±4.45 | 0.358±0.259 | 0.381±0.112 | 0.123±0.007 | 0.0012±0.0007 | 0.022±0.021 |
| Gaussian Noise | 0.5 | 5 | 69.04±3.49 | 0.560±0.307 | 0.482±0.231 | 0.125±0.005 | 0.0010±0.0005 | 0.011±0.015 |
| DTD Textures | 0.0 | 5 | 71.34±4.31 | 0.433±0.242 | 0.143±0.081 | 0.129±0.028 | 0.0005±0.0006 | 0.000±0.000 |
| DTD Textures | 0.1 | 5 | 74.70±4.69 | 0.366±0.188 | 0.110±0.072 | 0.124±0.018 | 0.0000±0.0010 | 0.037±0.005 |

### B. CIFAR-100 Contamination Sweep (FixMatch, SVHN)

Table A2 extends Table II by adding PLE, Sem-Drift, Grad-Align, and OOD-FF alongside the best accuracy and CCI columns already reported for the CIFAR-100 / FixMatch / SVHN sweep.

TABLE A2
FIXMATCH ON CIFAR-100 WITH SVHN CONTAMINATION: FULL METRIC SWEEP

| r | n | Acc. (%) | PLE | CCI | Sem-Drift | Grad-Align | OOD-FF |
|---|---|---|---|---|---|---|---|
| 0.0 | 5 | 43.21±1.20 | 1.827±0.109 | 0.148±0.022 | 0.390±0.009 | 0.0004±0.0002 | 0.000±0.000 |
| 0.1 | 5 | 43.47±1.35 | 1.424±0.162 | 0.321±0.031 | 0.413±0.011 | 0.0003±0.0001 | 0.061±0.022 |
| 0.2 | 5 | 43.51±1.51 | 1.303±0.236 | 0.551±0.095 | 0.442±0.017 | 0.0002±0.0002 | 0.059±0.010 |
| 0.3 | 5 | 42.47±1.13 | 1.435±0.217 | 0.694±0.145 | 0.431±0.003 | 0.0004±0.0001 | 0.038±0.012 |
| 0.4 | 5 | 41.63±1.60 | 1.215±0.170 | 1.153±0.125 | 0.434±0.007 | 0.0004±0.0001 | 0.054±0.011 |
| 0.5 | 5 | 39.22±1.40 | 1.321±0.375 | 1.383±0.274 | 0.439±0.022 | 0.0004±0.0003 | 0.049±0.020 |

### C. Algorithm Comparison Under SVHN Contamination (CIFAR-10)

Table A3 extends the comparison in Fig. 3 by reporting all six metrics for FixMatch, FlexMatch, SoftMatch, and DS3L under SVHN contamination on CIFAR-10, across the full contamination-ratio sweep $r \in \{0.0, ..., 0.5\}$, where $r = 0.0$ denotes the clean (no-OOD) baseline for each algorithm.

TABLE A3
ALGORITHM COMPARISON ON CIFAR-10 WITH SVHN CONTAMINATION: FULL METRIC SWEEP

| Algorithm | r | n | Acc. (%) | PLE | CCI | Sem-Drift | Grad-Align | OOD-FF |
|---|---|---|---|---|---|---|---|---|
| FixMatch | 0.1 | 5 | 74.55±3.90 | 0.252±0.182 | 0.104±0.055 | 0.113±0.020 | 0.0009±0.0012 | 0.042±0.009 |
| FixMatch | 0.2 | 5 | 72.60±5.41 | 0.151±0.033 | 0.225±0.042 | 0.106±0.004 | 0.0015±0.0014 | 0.037±0.002 |
| FixMatch | 0.3 | 5 | 73.40±5.95 | 0.169±0.065 | 0.329±0.061 | 0.113±0.004 | 0.0005±0.0010 | 0.037±0.006 |
| FixMatch | 0.4 | 5 | 72.23±3.71 | 0.336±0.271 | 0.377±0.144 | 0.123±0.011 | 0.0003±0.0003 | 0.027±0.013 |
| FixMatch | 0.5 | 5 | 70.51±5.60 | 0.393±0.331 | 0.552±0.173 | 0.131±0.011 | 0.0005±0.0004 | 0.028±0.018 |
| FlexMatch | 0.0 | 5 | 79.19±2.74 | 0.366±0.172 | 0.018±0.007 | 0.153±0.027 | 0.0016±0.0008 | 0.000±0.000 |
| FlexMatch | 0.1 | 5 | 79.05±4.97 | 0.244±0.054 | 0.018±0.006 | 0.133±0.015 | 0.0009±0.0005 | 0.033±0.001 |
| FlexMatch | 0.2 | 5 | 73.36±8.51 | 0.204±0.044 | 0.050±0.030 | 0.141±0.007 | 0.0011±0.0006 | 0.030±0.001 |
| FlexMatch | 0.3 | 5 | 69.11±4.14 | 0.256±0.078 | 0.087±0.031 | 0.152±0.005 | 0.0005±0.0005 | 0.028±0.001 |
| FlexMatch | 0.4 | 5 | 68.57±6.16 | 0.257±0.076 | 0.179±0.035 | 0.162±0.010 | 0.0008±0.0007 | 0.027±0.002 |
| FlexMatch | 0.5 | 5 | 69.43±4.36 | 0.363±0.128 | 0.285±0.056 | 0.165±0.007 | 0.0007±0.0014 | 0.027±0.002 |
| SoftMatch | 0.0 | 5 | 72.65±2.59 | 0.822±0.165 | 0.018±0.004 | 0.242±0.025 | -0.0000±0.0007 | 0.000±0.000 |
| SoftMatch | 0.1 | 5 | 72.36±5.31 | 0.671±0.243 | 0.035±0.002 | 0.218±0.036 | 0.0003±0.0011 | 0.032±0.001 |
| SoftMatch | 0.2 | 5 | 76.54±2.99 | 0.569±0.143 | 0.057±0.010 | 0.202±0.017 | 0.0007±0.0007 | 0.032±0.001 |
| SoftMatch | 0.3 | 5 | 73.24±5.96 | 0.585±0.324 | 0.104±0.051 | 0.195±0.049 | 0.0009±0.0009 | 0.032±0.001 |
| SoftMatch | 0.4 | 5 | 72.76±3.17 | 0.613±0.185 | 0.148±0.028 | 0.194±0.015 | 0.0011±0.0009 | 0.031±0.002 |
| SoftMatch | 0.5 | 5 | 68.75±4.63 | 0.489±0.184 | 0.276±0.084 | 0.190±0.014 | 0.0004±0.0008 | 0.032±0.001 |
| DS3L | 0.0 | 5 | 72.48±6.47 | 0.500±0.212 | 0.121±0.070 | 0.141±0.014 | 0.0007±0.0007 | 0.000±0.000 |
| DS3L | 0.1 | 5 | 75.87±3.92 | 0.282±0.194 | 0.120±0.061 | 0.117±0.020 | 0.0012±0.0009 | 0.045±0.012 |
| DS3L | 0.2 | 5 | 71.99±7.17 | 0.204±0.059 | 0.211±0.044 | 0.111±0.008 | 0.0012±0.0005 | 0.041±0.006 |
| DS3L | 0.3 | 5 | 73.32±3.55 | 0.205±0.091 | 0.310±0.028 | 0.114±0.009 | 0.0009±0.0006 | 0.040±0.007 |
| DS3L | 0.4 | 5 | 70.71±4.52 | 0.204±0.096 | 0.455±0.101 | 0.121±0.006 | 0.0013±0.0011 | 0.039±0.005 |
| DS3L | 0.5 | 5 | 70.17±2.43 | 0.282±0.198 | 0.613±0.099 | 0.122±0.006 | 0.0005±0.0011 | 0.034±0.015 |